\documentclass[preprint,12pt, a4paper]{elsarticle}

\usepackage{amssymb}
\usepackage{hyperref}
\usepackage{listings}
\usepackage{color}

\definecolor{dkgreen}{rgb}{0,0.6,0}
\definecolor{gray}{rgb}{0.5,0.5,0.5}
\definecolor{mauve}{rgb}{0.58,0,0.82}

\journal{SoftwareX}

\begin{document}
\renewcommand{\labelenumii}{\arabic{enumi}.\arabic{enumii}}

\begin{frontmatter}
\title{yProv4ML: Effortless Provenance Tracking\\ for Machine Learning Systems}


\author[a1]{G. Padovani}
\author[a2]{V. Anantharaj}
\author[a1]{S. Fiore}
\address[a1]{University of Trento, Italy}
\address[a2]{Oak Ridge National Laboratory, USA}

\begin{abstract}


The rapid growth of interest in large language models (LLMs) reflects their potential for flexibility and generalization, and attracted the attention of a diverse range of researchers. However, the advent of these techniques has also brought to light the lack of transparency and rigor with which development is pursued. In particular, the inability to determine the number of epochs and other hyperparameters in advance presents challenges in identifying the best model. 
To address this challenge, machine learning frameworks such as MLFlow can automate the collection of this type of information. However, these tools capture data using proprietary formats and pose little attention to lineage. This paper proposes yProv4ML, a framework to capture provenance information generated during machine learning processes in PROV-JSON format, with minimal code modifications.
\end{abstract}


\end{frontmatter}



\begin{table}[!h]
\begin{tabular}{|l|p{6.5cm}|p{6.5cm}|}
\hline
\textbf{Nr.} & \textbf{Code metadata description} & \textbf{Please fill in this column} \\
\hline
C1 & Current code version & v1.0 \\
\hline
C2 & Permanent link to code/repository used for this code version & \url{https://github.com/HPCI-Lab/yProvML} \\
\hline
C3  & Permanent link to Reproducible Capsule & \\
\hline
C4 & Legal Code License   & GPLv3 \\
\hline
C5 & Code versioning system used & git \\
\hline
C6 & Software code languages, tools, and services used & Python \\
\hline
C7 & Compilation requirements, operating environments \& dependencies & Codecarbon, Prov, Pytorch \\
\hline
C8 & If available Link to developer documentation/manual & \url{https://hpci-lab.github.io/yProv4ML.github.io/} \\
\hline
C9 & Support email for questions & \textit{gabriele.padovani@unitn.it} \\
\hline
\end{tabular}
\caption{Code metadata}
\label{codeMetadata} 
\end{table}

\section{Motivation and significance}

The field of machine learning has experienced a remarkable acceleration in recent years, with new findings being superceded within weeks. While this rapid pace of research undoubtedly offers numerous benefits, it has also led to a prevalence of works conducted with less rigor and in a superficial way. 
Code that is not accompanied by documentation and results that are not reproducible inevitably lead to confusion among researchers and an environment in which trust is not a fundamental aspect of the proposed work \cite{NONREPROD}. 

The complexity of the data manipulation process, which frequently involves ad hoc and repeated transformations, further complicates matters. Several issues may arise when attempting to reproduce these steps in order to trace the creation process. In such instances, the transformation may be lost, resulting in a slight difference between the original design process and the finalized one \cite{REPEAT}.
It is becoming increasingly important to meticulously document the entirety of the design and development process, in order to facilitate comprehensive replication of experiments and prevent the introduction of unverified outcomes. 

In addition to these difficulties, it is often computationally expensive to determine the value of numerous hyperparameters employed during the training of machine learning models. The most common approach is to perform multiple tests to identify the parameter that performs optimally \cite{REPROD}. With an historical record of experiments, users could look up similar targets, and identify hyperparameter values which could be ideal for their application. 

A further issue with hyperparameter tuning arises from repeated attempts at training the optimal model. When the same process is iterated several times, a considerable amount of computing resources are wasted unnecessarily. Given the large size of many machine learning models, this paradigm quickly becomes unsustainable, particularly when dealing with deep learning architectures comprising billions of parameters. 
To solve this issue, some approaches try to solve the parameter optimization problem during the model training phase, in an online manner \cite{DEEPHYPER}, though this knowledge is, currently, not easily reusable (transferable) to similar experiments run by other researchers.

In this context, provenance — \textit{the record of the origins, history, and transformations of data, models, and decisions}— \cite{cheney2009provenance} plays a crucial role. 
While similar approaches have been developed, both in the context of climate science and HPC domain \cite{SOUZA2}, with the aim of tracking the entire workflow lineage \cite{SOUZA1}, this work focuses on the specific ML task, allowing for more fine-grained lineage collection. 
yProv4ML offers an accessible and intuitive approach to storing information about the dataset, hyperparameters, and energy efficiency metrics. Its functionalities can be accessed in a manner analogous to MLFlow \cite{MLFLOW}, facilitating a seamless transition. Furthermore, all content that the user elects to track will be stored in PROV-JSON \cite{PROVJSON}, which has become the established standard for recording provenance artifacts.  

\section{Software description}

We have developed the yProv4ML library that provides access to MLFlow-analogous logging utilities, offering a recognizable interface for the collection and storage of provenance data, and gathers three primary categories of information: artifacts, parameters, and metrics. 
The former identifies any file or output utilized in subsequent phases of the workflow. In the context of machine learning processes, these predominantly encompass model versions, checkpoints, and source code material. 
In contrast, parameters represent one-time logged values utilized during the training phase. Examples of such values include the learning rate, the size and width of the model, and other hyper-parameters. 
The final category contains information that is updated during the training process. This includes metrics such as loss values and statistics related to the execution of the program, such as energy efficiency, power consumption, and GPU usage. 

Once data recorded during a single run has been stored, the library enables a comparison between the results of consecutive related executions. This facilitates a more comprehensive understanding of the influence of hyper-parameters and model configurations, while simultaneously maintaining an accurate record of any alterations made to the entire script. 

\begin{figure}
    \centering
\includegraphics[width=0.8\textwidth]{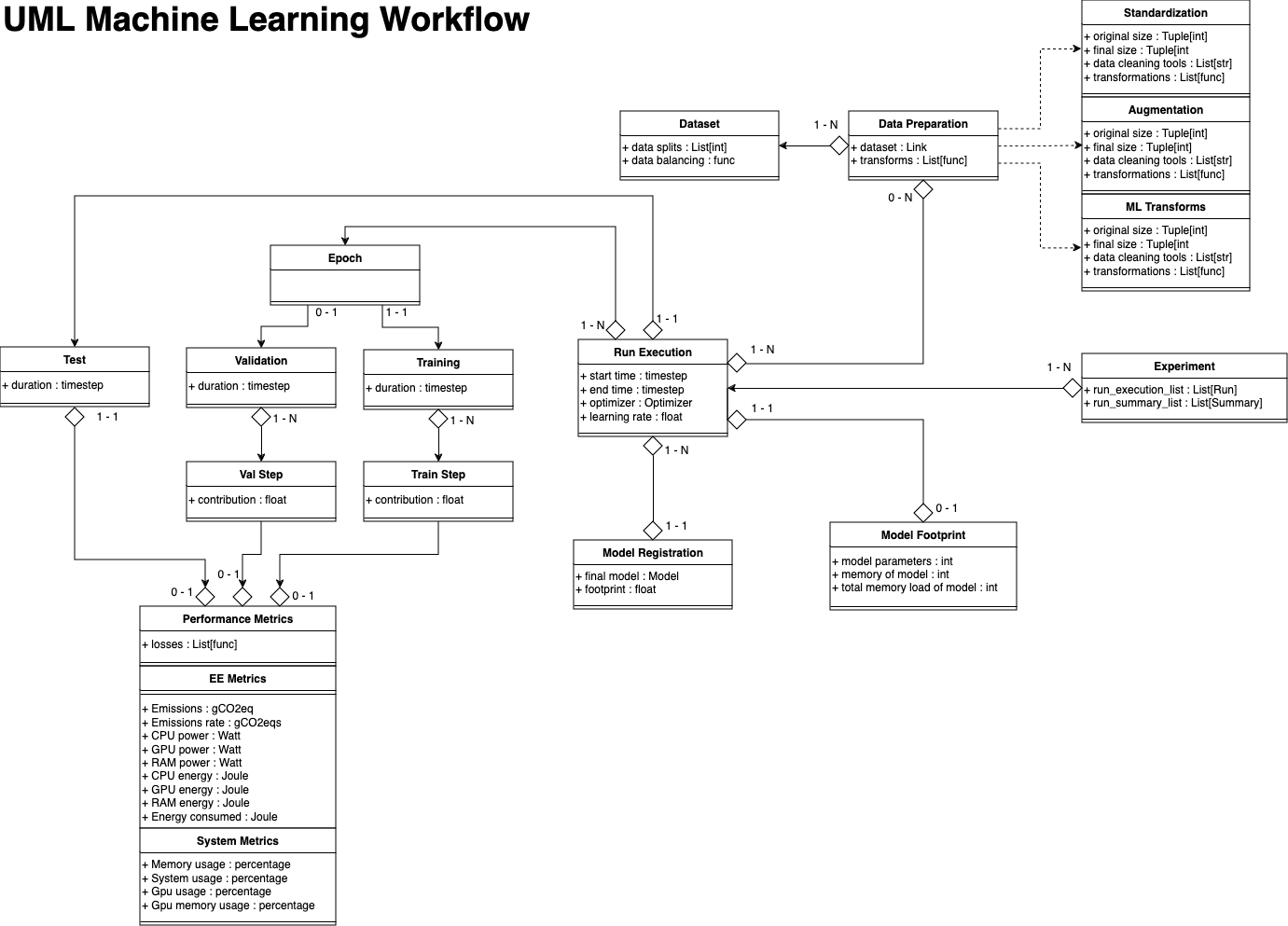}
    \caption{Data Model used as foundation for yProv4ML}
    \label{figure:data_model}
\end{figure}

\subsection{Software architecture}

yProv4ML employs an ad hoc data model that enables the efficient encapsulation of all data collected during program execution within a reduced memory footprint. 
The data model used in the library is shown in Fig. \ref{figure:data_model}. 

The main modules which make up the core components are four: 
\begin{itemize}
    \item \textbf{Main module}: which wraps all functionality of the library and allows for context declaration and shutdown. It also exposes several directives to allow the user to log information and customize the yProv4ML experience; 
    \item \textbf{Energy module}: it contains all utility functions to save energy-related metrics, such as emissions, power consumption and others; 
    \item \textbf{System module}: contains directives to save information related to the system, such as memory or GPU usage; 
    \item \textbf{Time module}: it contains helper functions to manipulate and save information. 
\end{itemize}

In addition to the modules mentioned above, the library is also capable of constructing a provenance graph, which contains all the data saved during the execution of the program. This graph encompasses all data logged by the user, as well as a set of automatically saved information, such as environment variables, required libraries, and others. 
Moreover, the generated graph is fully interoperable with other libraries of the yProv framework \cite{YPROV}, such as yProv4WFs. In the event that the user elects to utilize both libraries, the output will exhibit a hierarchical provenance graph, comprising the workflow information at the higher level and the ML task information at a lower more detailed level.

 \subsection{Software functionalities}

The primary objective of yProv4ML is to facilitate the extraction of provenance information in an intuitive manner. To achieve this, the directives accessible through this library were designed to draw upon established tools such as MLFlow, thereby enabling a seamless transition between the two. 
The execution of any experiment starts with a call to the \texttt{start\_run} function and concludes with a call to \texttt{end\_run}. The following sections provide a detailed explanation of these directives. 

\begin{lstlisting}
prov4ml.start_run(
    prov_user_namespace: str,
    experiment_name: Optional[str] = None,
    provenance_save_dir: Optional[str] = None,
    collect_all_processes: Optional[bool] = False,
    save_after_n_logs: Optional[int] = 100,
    rank : Optional[int] = None, 
)
# code...
prov4ml.end_run(
    create_graph: Optional[bool] = False, 
    create_svg: Optional[bool] = False, 
)
\end{lstlisting}

This section will present an overview of the primary functions utilized for the logging of metrics and parameters within the ProvJSON file. A subset of these functions is provided below as reference. 

\begin{lstlisting}
prov4ml.log_model(model_label, model)
\end{lstlisting}

The directive \texttt{log\_model} is employed for the purpose of recording information about the architecture configuration trained during the course of the experiment. 
The majority of the metrics saved during this phase pertain to the model's memory footprint, including the number of parameters, the memory allocation utilized by the network, and the gradients. Additionally, the complete composition of the model can also be logged in the provenance file, which includes all layers structure, data types, and input and output size.  

\begin{lstlisting}
prov4ml.log_param(key, value)
\end{lstlisting}

The function \texttt{log\_param} enables the logging of a single parameter key-value pair, which can be stored either in program memory or in temporary files. Consequently, any given value can be logged only once with that specific key. 

\begin{lstlisting}
prov4ml.log_metric(key, value, context, step)
\end{lstlisting}

In contrast, the function \texttt{log\_metric} is employed when a sequence of parameters must be stored. This is typically utilized for logging losses or other analogous metrics that fluctuate over time. 

As previously stated, the context parameter is utilized to assign a specific level of importance to the metric, thereby facilitating both grouping and the construction of a hierarchical data structure. The available contexts are training, evaluation, and validation; however, additional contexts may be declared and employed in a customized manner. 
The step parameter, on the other hand, indicates the timestamp at which the metric value will be saved, thus enabling the tracking of its behavior during the learning process. 

\begin{lstlisting}
prov4ml.log_system_metrics(context, step)
\end{lstlisting}

The directive \texttt{log\_system\_metrics} is employed to record data pertaining to the memory and resources utilized by the entire program. The statistics encompass total memory usage, disk usage, GPU memory usage, and GPU usage. 

\begin{lstlisting}
prov4ml.log_carbon_metrics(context, step)
\end{lstlisting}

Similarly to the previous function, \texttt{log\_carbon\_metrics} is used to save information about system metrics from the point of view of energy efficiency. These metrics include emissions, CPU and GPU power consumed, as well as other memory statistics. 

\begin{lstlisting}
prov4ml.log_artifact(artifact_path, context, step, timestamp)
\end{lstlisting}

The function \texttt{log\_artifact} is employed for the purpose of saving data in file format, in particular for information utilized or generated during the course of program execution. In addition, the directive also allows to specify the context, timestamp and the step parameter, to inform the reader of and potential modifications to these documents.  

\begin{lstlisting}
prov4ml.save_model_version(model,model_name, context, step, timestamp)
\end{lstlisting}

The \texttt{save\_model\_version} directive is utilized for the purpose of saving a checkpoint of the model in PyTorch format, accompanied by the documentation of pertinent information regarding the file in question, which is then saved as an artifact. 

\begin{lstlisting}
prov4ml.log_current_execution_time(time_label, context, step)
\end{lstlisting}

At last, the \texttt{log\_current\_execution\_time} function may be employed to assign a timestamp to a designated label. This function offers a straightforward interface for recording the time required to complete specific operations, and the resulting data can be plotted in a manner analogous to any other metric. 
  
\subsection{Sample code snippets analysis} 

The following code snippet illustrates a potential application of the yProv4ML library with PyTorch, and can be additionally integrated with PyTorch Lightning loggers. 
The \texttt{prov4ml. start\_run} function is invoked to initialize the context and to maintain a record of the total execution time.

The user can then elect to record artifacts, metrics and parameters coming from the training execution, and all will be stored inside the final PROV-JSON file. 

Ultimately, the context is terminated and the PROV-JSON file is stored in the designated directory via the invocation of the \texttt{prov4ml.end\_run} function. Subsequent to the file's saving, the user is allowed to transform the JSON file into an SVG image, while the identical information in DOT format is automatically saved. 

\begin{lstlisting}
# start the run in the same way as with mlflow
prov4ml.start_run(
    prov_user_namespace="www.example.org",
    experiment_name="experiment_name", 
    provenance_save_dir="prov",
    save_after_n_logs=100,
)
    
mnist_model = MNISTModel()
# log the dataset transformation as one-time parameter
tform = transforms.Compose([RandomRotation(10), ToTensor()])
prov4ml.log_param("dataset transformation", tform)

train_ds = MNIST(PATH_DATASETS, transform=tform)
train_loader = DataLoader(train_ds, batch_size=BATCH_SIZE)
prov4ml.log_dataset(train_loader, "train_dataset")

for epoch in tqdm(range(EPOCHS)):
    mnist_model.train()
    for i, (x, y) in enumerate(train_loader):
        optim.zero_grad()
        y_hat = mnist_model(x)
        loss = loss_fn(y_hat, y)
        loss.backward()
        optim.step()
    
        # log system and carbon metrics (once per epoch), as well as the execution time
        prov4ml.log_metric("MSE_train", loss.item(), context=prov4ml.Context.TRAINING, step=epoch)
        prov4ml.log_carbon_metrics(prov4ml.Context.TRAINING, step=epoch)
        prov4ml.log_system_metrics(prov4ml.Context.TRAINING, step=epoch)
    # save incremental model versions
    prov4ml.save_model_version(mnist_model, "mnist_model_version", prov4ml.Context.TRAINING, epoch)

prov4ml.log_model(mnist_model, "mnist_model_final")
# save the provenance graph
prov4ml.end_run(create_graph=True, create_svg=True)
\end{lstlisting}

Once the function \texttt{prov4ml.end\_run} is invoked, each process generates its own provenance file, which will contain only the data pertinent to its execution. Furthermore, all PROV-JSON files can be linked together using an additional PROV collection, which serves as a summary of the distributed execution.  

\section{Illustrative examples}

\begin{figure}[t!]
    \centering
    \includegraphics[height=\textheight]{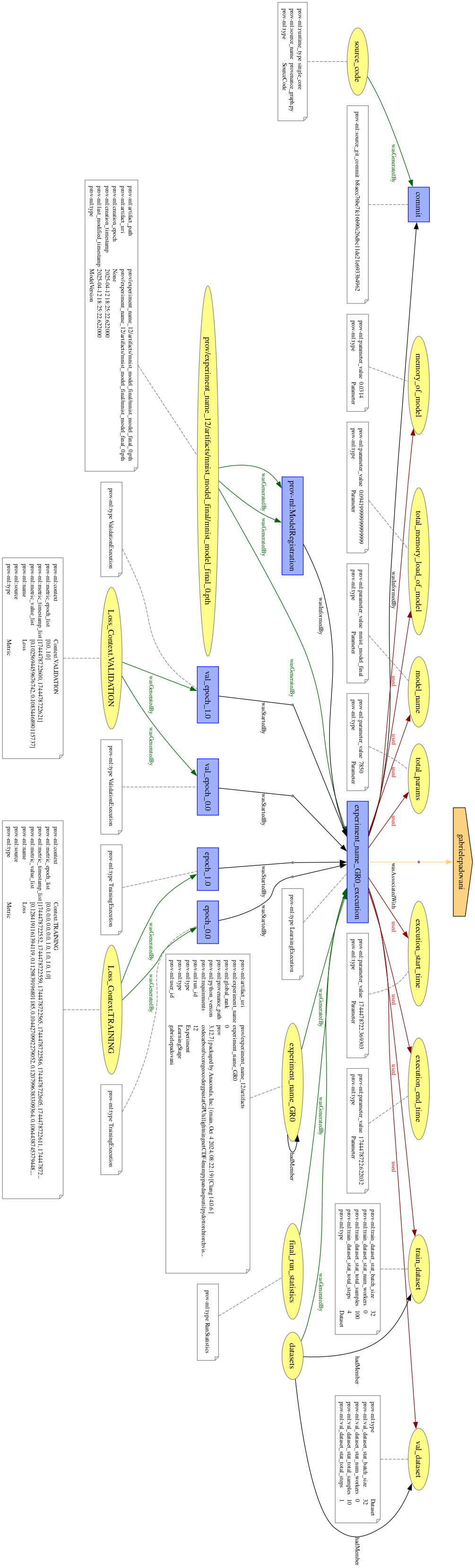}
    \caption{Provenance graph of the ML training process specified in Section 3 (Illustrative examples). The figure is available in svg format at \texttt{\url{https://github.com/HPCI-Lab/yProvML}}}
    \label{figure:example}
\end{figure}

An example of a provenance graph obtainable using yProv4ML is shown in Fig.\ref{figure:example}. 
In this use case the machine learning application is a simple MNIST \cite{deng2012mnist} classification task created with the code snippet shown above. It makes use of a very small network and a single epoch of training to reach acceptable performance levels, but different experiments could be recorded, and larger provenance graphs could be constructed. 

\begin{figure}[tb]
    \centering
    \includegraphics[width=\textwidth]{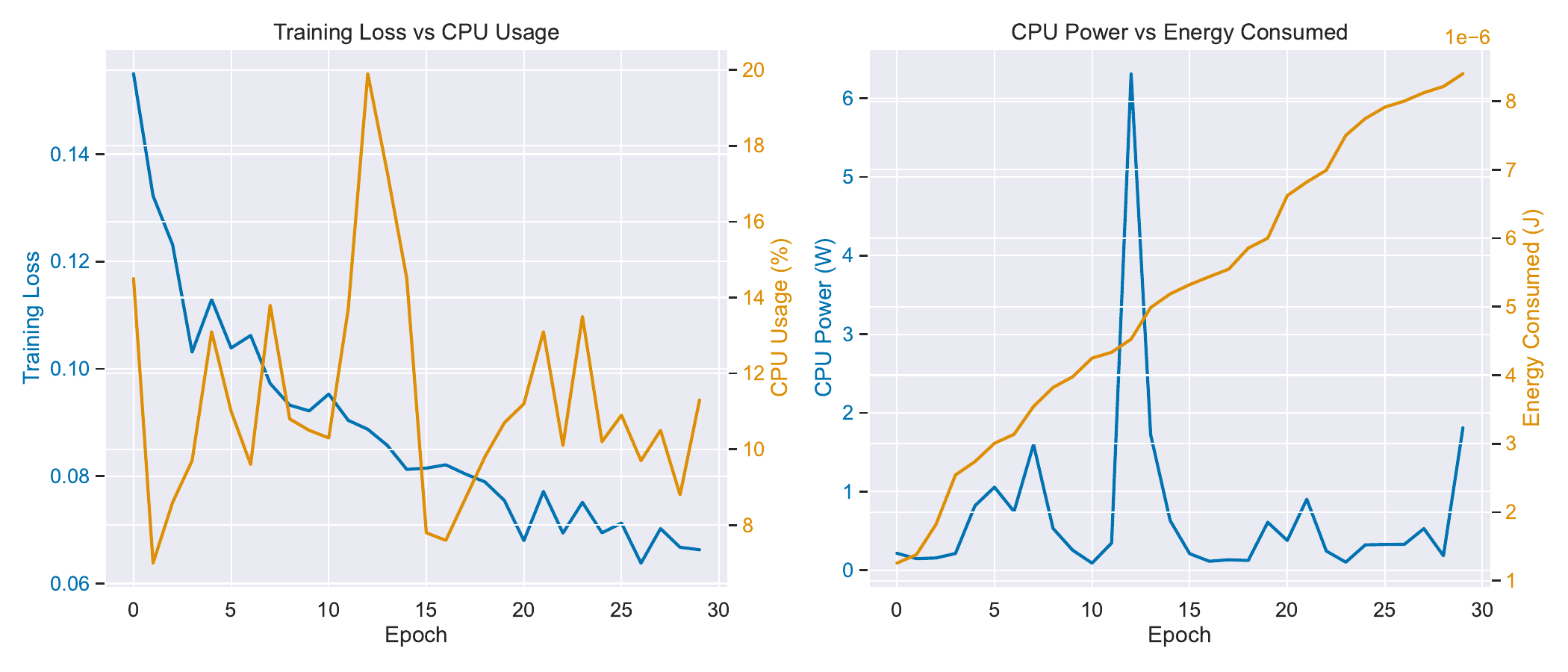}

    \caption{Left: the training loss is recorded for every epoch using yProv4ML directives and compared to the CPU usage during training. Right: CPU power consumption is recorded for every batch and compared with the total energy consumed during the same training process.}
    \label{figure:train_loss}
\end{figure}

In Fig. \ref{figure:train_loss} two possible metrics are represented and saved using yProv4ML. The metrics are put in relation according to the current logging time, and the user is able to decide how to collect, process and compare the data recorded.
The figure on the left shows CPU usage spikes with relation to the loss achieved by the model at the current epoch. In both cases, data is collected after each sample passes through the model. 
In the figure on the right, on the other hand, CPU power consumption is recorded for each sample, as well as the total energy consumed by the training, calculated in real time during the run. 

\section{Impact}

The usefulness of yProv4ML is evident in the context of monitoring energy consumption processes. A preliminary testing phase can be employed to understand which component of the program exerts the greatest influence, thereby facilitating a reduction in overall consumption during the subsequent training phase. 
The ability to log several versions of the same experiment is also crucial for understanding which hyper-parameters work better with the current execution, and avoiding repeating the same mistakes over several runs. 

yProv4ML has been employed to assess the outcome of a series of benchmarking experiments, as outlined in \cite{VAL}, conducted in collaboration with the Oak Ridge National Laboratory (ORNL). In this context, the library was evaluated on single-node, multi-node programs, as well as on over than 500 GPUs in parallel. 
The benchmarks were designed to understand the quantity of computation, data, and parameters required for optimal training of a series of foundation models. In this context, the capacity to readily save a substantial number of provenance artifacts has proven to be of great importance. 

Similarly, yProv4ML was employed to gather data on a machine learning process designed to predict the occurrence and behavior of tropical cyclones \cite{MAX}. In this instance, the library was used to log information about the graph neural network which was being developed. 




\section{Conclusions}

In this paper, we propose yProv4ML, a library for serialization in JSON format of provenance information collected during a machine learning process, in accordance with the W3C PROV standard. It provides the user with a transparent and user-friendly tool capable of translating the collected information into the standard format defined by PROV-JSON. Users who adopt yProv4ML are not restricted to proprietary options when having to visualize or edit metadata from machine learning processes, and information is saved in JSON format, which is also human readable. 

In subsequent development, the emphasis will be on the aggregation of more comprehensive provenance data. This may encompass the delineation of the contribution of each sample, the monitoring of the impact of varying parameters, and the identification of the most suitable parameters in view of specific metrics. 


\section*{Acknowledgements}

\textit{This work was partially funded under the National Recovery and Resilience Plan (NRRP), Mission 4 Component 2 Investment 1.4 - Call for tender No. 1031 of 17/06/2022 of Italian Ministry for University and Research funded by the European Union – NextGenerationEU (proj. nr. CN\_00000013) and the EU InterTwin project (Grant Agreement 101058386).}

\textit{Moreover this research used resources of the Oak Ridge Leadership Computing Facility at the Oak Ridge National Laboratory, which is supported by the Office of Science of the U.S. Department of Energy under Contract No. DE-AC05-00OR22725.}



\bibliographystyle{IEEEtran}
\bibliography{refs}

\end{document}